\documentclass[conference, 10 pt]{IEEEtran}  

\IEEEoverridecommandlockouts

\usepackage{cite}
\usepackage{amsmath,amssymb,amsfonts}
\usepackage{algorithmic}
\usepackage{graphicx}
\usepackage{textcomp}
\usepackage[dvipsnames]{xcolor}
\usepackage{geometry}
\usepackage{booktabs}
\usepackage{ulem}
\def\BibTeX{{\rm B\kern-.05em{\sc i\kern-.025em b}\kern-.08em
    T\kern-.1667em\lower.7ex\hbox{E}\kern-.125emX}}

\usepackage{scalerel}
\usepackage{tikz}
\usetikzlibrary{svg.path}

\definecolor{orcidlogocol}{HTML}{A6CE39}
\tikzset{
  orcidlogo/.pic={
    \fill[orcidlogocol] svg{M256,128c0,70.7-57.3,128-128,128C57.3,256,0,198.7,0,128C0,57.3,57.3,0,128,0C198.7,0,256,57.3,256,128z};
    \fill[white] svg{M86.3,186.2H70.9V79.1h15.4v48.4V186.2z}
                 svg{M108.9,79.1h41.6c39.6,0,57,28.3,57,53.6c0,27.5-21.5,53.6-56.8,53.6h-41.8V79.1z M124.3,172.4h24.5c34.9,0,42.9-26.5,42.9-39.7c0-21.5-13.7-39.7-43.7-39.7h-23.7V172.4z}
                 svg{M88.7,56.8c0,5.5-4.5,10.1-10.1,10.1c-5.6,0-10.1-4.6-10.1-10.1c0-5.6,4.5-10.1,10.1-10.1C84.2,46.7,88.7,51.3,88.7,56.8z};
  }
}

\newcommand\orcidicon[1]{\href{https://orcid.org/#1}{\mbox{\scalerel*{
\begin{tikzpicture}[yscale=-1,transform shape]
\pic{orcidlogo};
\end{tikzpicture}
}{|}}}}

\usepackage{hyperref} 

\usepackage{atbegshi}
\usepackage[absolute,overlay]{textpos}
\usepackage{rotating}
\usepackage[english]{babel}
\usepackage{blindtext}

\usepackage{framed}

\begin{document}

\newcommand\mc[1]{\multicolumn{1}{c}{#1}} 




\title{Shadow Erosion and Nighttime Adaptability for Camera-Based Automated Driving Applications}

\author{Mohamed Sabry\orcidicon{0000-0002-9721-6291} \textit{Member, IEEE}, Gregory Schroeder\orcidicon{0009-0005-7340-1715} \textit{Member, IEEE}, \\ Joshua Varughese\orcidicon{0000-0002-3250-0742} \textit{Member, IEEE} and Cristina Olaverri-Monreal\orcidicon{0000-0002-5211-3598} \textit{Senior Member, IEEE}%
\thanks{ Johannes Kepler University Linz, Austria, Department Intelligent
Transport Systems, Altenberger Straße 69, 4040 Linz, Austria.
\texttt{\{mohamed.sabry, gregory.schroeder, joshua.varughese, cristina.olaverri-monreal\}@jku.at}}%
}

\maketitle
\thispagestyle{empty}
\pagestyle{empty}
\setlength{\textfloatsep}{1\baselineskip plus 0.2\baselineskip minus 0.2\baselineskip}

\setlength{\dbltextfloatsep}{1\baselineskip plus 0.2\baselineskip minus 0.2\baselineskip}
\begin{abstract}

Enhancement of images from RGB cameras is of particular interest due to its wide range of ever-increasing applications such as medical imaging, satellite imaging, automated driving, etc. 
In autonomous driving, various techniques are used to enhance image quality under challenging lighting conditions. These include artificial augmentation to improve visibility in poor nighttime conditions, illumination-invariant imaging to reduce the impact of lighting variations, and shadow mitigation to ensure consistent image clarity in bright daylight. This paper proposes a pipeline for Shadow Erosion and Nighttime Adaptability in images for automated driving applications while preserving color and texture details. The Shadow Erosion and Nighttime Adaptability pipeline is compared to the widely used CLAHE technique and evaluated based on illumination uniformity and visual perception quality metrics. The results also demonstrate a significant improvement over CLAHE, enhancing a YOLO-based drivable area segmentation algorithm.

\end{abstract}

\begin{textblock*}{18.15cm}(1.55cm,26cm) 
\begin{minipage}{17.8cm}
     \vspace{0.1cm} 
     {\footnotesize\copyright 2025 IEEE. Personal use of this material is permitted. Permission from IEEE must be obtained for all other uses, in any current or future media, including reprinting/republishing this material for advertising or promotional purposes, creating new collective works, for resale or redistribution to servers or lists, or reuse of any copyrighted component of this work in other works. DOI: 10.1109/IV64158.2025.11097428}
\end{minipage}
\end{textblock*}

%
\section{Introduction}
\label{sec:Introduction}
Image enhancement plays a crucial role in Automated Driving (AD). In AD systems, image quality critically impacts the performance of camera-based algorithms for perception, localization, decision-making \cite{guo2016lime} etc. In the particular case of Platooning, advanced perception systems utilizing cameras help to maintain safe and efficient vehicle coordination. However, challenging lighting conditions, such as low light, glare, fog, or shadows, can degrade image quality, leading to unreliable object detection, lane recognition, and drivable area segmentation.
Therefore, pre-processing is essential to address challenges such as shadows in sunny daytime images and poor illumination at night, which could possibly degrade system performance.

Camera-based algorithms often struggle under challenging lighting conditions, requiring transformations to improve results. \cite{maddern2014illumination} demonstrated that illumination-invariant images enhance applications like localization and segmentation. Beyond algorithmic improvements, AD systems at SAE Level 4 and above also require images suitable for human interpretation during remote monitoring \cite{wenliang2024research}, making image enhancement vital while preserving human-perceivable details. Existing methods typically focus on either nighttime illumination enhancement or on daytime shadow reduction \cite{mei2024latent}. Illumination-invariant color spaces \cite{gevers1999color, maddern2014illumination} can handle both, but often result in color loss and inconsistencies across cameras due to parameter dependencies. In addition, previous correction methods applied to wide-view images, such as those captured from vehicle dashboard cameras, have revealed the presence of varying shadow intensities. Compensating for all such shadows within an image poses significant challenges, primarily due to pronounced illumination variations, which can significantly influence the chroma channels and hinder effective shadow correction. Inadequate contrast in the AD context can also lead to critical situations \cite{certad2024inadequate}.

This paper presents a real-time image enhancement method for shadow mitigation in sunny daytime conditions and improved nighttime visibility. Unlike prior methods, it does not require hardware calibration or smoothing algorithms, preserving image details and texture. The method is evaluated against CLAHE \cite{zuiderveld1994contrast} using six Blind Image Quality Assessment (BIQE) metrics. 
Additionally, a YOLO-based Deep Learning (DL) model for drivable area segmentation was utilized with the proposed method as a preprocessing step, demonstrating its enhanced performance on nighttime images compared to CLAHE and emphasizing its potential for advanced Image Processing (IP) and Computer Vision (CV) tasks.


This paper is structured as follows: Section II reviews related work, Section III provides a detailed description of the Shadow Erosion and Nighttime Adaptability (SENA) pipeline presented in this work. Section IV presents the experimental setup, followed by Section V showing the results. Finally, Section VI summarizes the contributions of this work and outlines potential directions for future research.

\vspace{-10pt}
\begin{figure*}[htp!]
\centering
  \includegraphics[width=0.85\textwidth]{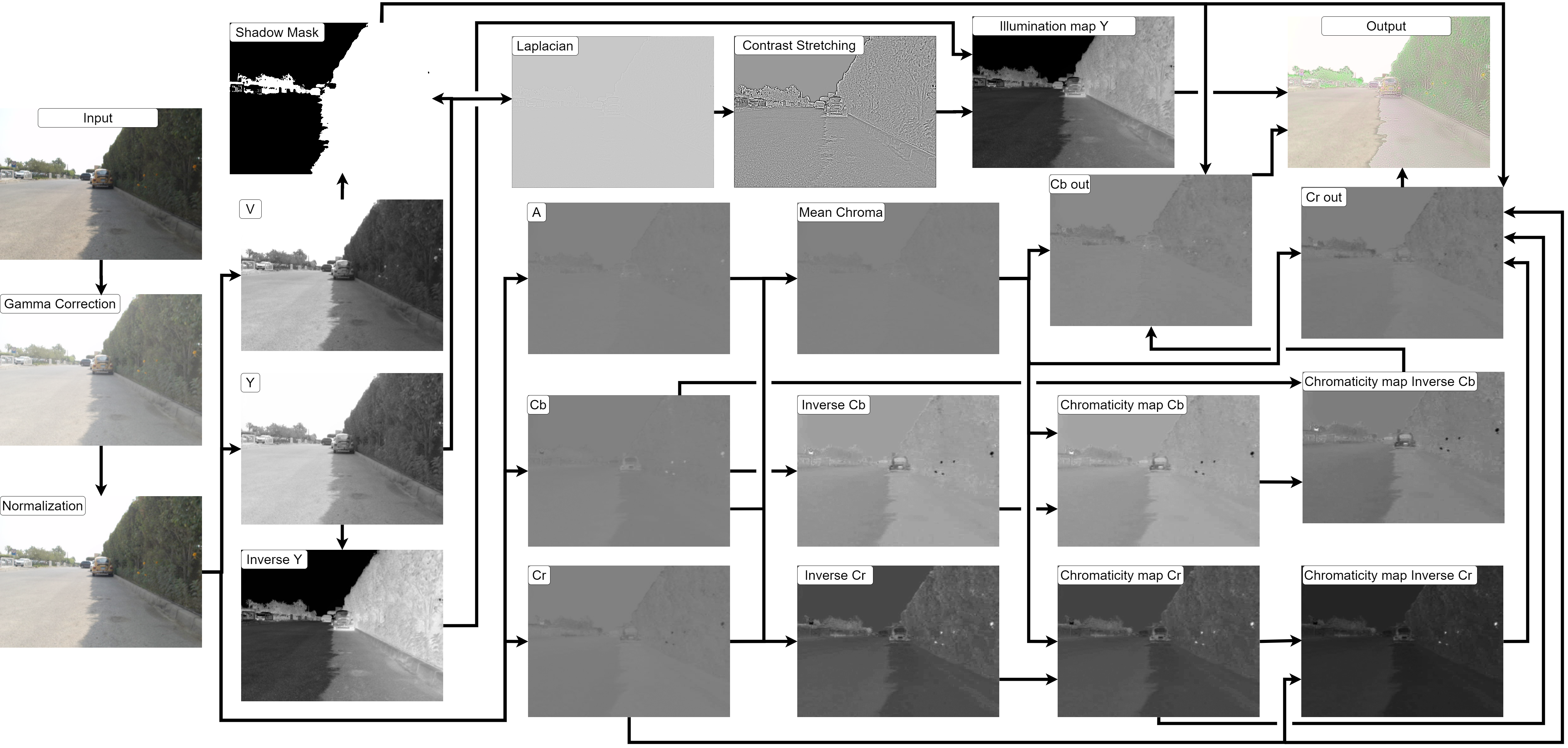}
  \caption{
  This figure illustrates the SENA pipeline steps from input to output.
  }
  \label{fig:follower_system}
\end{figure*}

\vspace{-1pt}

\section{Related Work}
\label{sec:RelatedWork} 

\vspace{-5pt}
Previous studies in image enhancement can be broadly categorized into three major directions: (1) generation of illumination-invariant images to mitigate the effects of illumination on various modules, (2) correction of poor and uneven illumination, and (3) shadow reduction in sunny daytime images.

\vspace{-5pt}

\subsection{Illumination-Invariant Images}

\vspace{-3pt}
The generation of illumination-invariant images aims to reduce or eliminate shadow effects, thereby improving the performance of vision-based systems. Illumination-invariant imaging has been demonstrated to improve modules in applications such as localization, mapping, and segmentation, particularly in autonomous driving scenarios \cite{maddern2014illumination}. For example, \cite{alvarez2010road} utilized illumination-invariant color spaces to segment drivable areas. Similarly, \cite{sabry2021drivable} incorporated illumination-invariant color spaces as a preprocessing step in drivable area segmentation modules. \cite{finlayson2005removal} employed logarithmic transformations to generate 2D log-chromaticity images. However, this approach requires camera-specific information to achieve optimal results.

\cite{shakeri2016illumination} enhanced this technique by introducing a Wiener filter based on the power law spectrum assumption of natural images, making the method more robust against illumination variations. Their approach also incorporated minimal entropy minimization between consecutive frames to reduce computational complexity.


Other researchers have applied illumination-invariant images for urban road segmentation by extracting road edges or enhancing segmentation performance under adverse weather conditions \cite{alshammari2019multi}. \cite{katramados2009real} employed a combination of IP and CV techniques, integrating multiple illumination-invariant methods, to accurately identify drivable areas for robotic navigation applications.

\subsection{Poor and Uneven Illumination Correction}

The challenge of correcting poor and uneven illumination has been approached using both traditional and DL based methods. Low-light images are characterized by reduced brightness and noise, necessitating preprocessing techniques such as gamma correction and the use of randomized parameters to adjust brightness levels. For instance, \cite{fu2024low} utilized low-illumination data preprocessing for criminal investigation tasks, though the method may not be suitable for AD scenarios.

Retinex-based algorithms, inspired by the Retinex theory of human visual perception \cite{jobson1997properties}, have been widely adopted for illumination correction. \cite{wang2019underexposed} proposed the DeepUPE network, which learns an image-to-illumination mapping by extracting global and local features. Enhanced images are obtained by dividing the input image by the illumination map, though the approach struggles with images containing predominantly dark regions or noise. \cite{seo2021deep} extended Retinex-based methods by employing image decomposition to improve detail visibility, while \cite{wang2021improved} refined the technique further with multiscale Retinex approaches. Other researchers used Wavelet Transform on satellite imaging to enhance image details \cite{elmasry2023image}.

Moreover, in \cite{tian2023survey}, a selection of DL approaches are presented, which use datasets with difficult poor illumination conditions, without focusing on shadows in sunny daytime driving.

In the context of autonomous driving, \cite{pham2020low} introduced DriveRetinex-Net, a Retinex-inspired network trained on a custom dataset. While effective, their results were primarily demonstrated on images captured near dusk or nighttime, excluding sunny daytime images with shadows. \cite{wei2018deep} proposed RetinexNet, a DL model based on Retinex theory, which enhances details in images. Modified Retinex approaches have also been employed for enhancing traffic visibility under nighttime conditions \cite{tao2022improved}.

\vspace{-5pt}

\subsection{Shadow Reduction in Daytime Images}

Shadow reduction in daytime images represents another crucial field in image enhancement research. For example, \cite{le2019shadow} developed a DL network based on shadow image decomposition, while \cite{saleh2024shadowremovalnet} proposed ShadowRemovalNet. Both methods demonstrated effectiveness on datasets containing diverse shadows at daytime but have not been specifically tailored for AD scenarios.

Previous studies have predominantly focused on either shadow reduction or enhancement of images under poor illumination conditions, with limited integration of both aspects. While illumination-invariant images address both challenges to some extent, they inherently lack color information and require camera-specific information and optimization for effective implementation. Furthermore, the adaptation of enhanced images specifically for AD applications has received minimal attention.

To address these gaps, this paper introduces SENA, a novel image enhancement pipeline tailored for AD applications. The proposed SENA approach combines advanced IP and CV techniques, including a modified illumination-invariant image method adapted from \cite{katramados2009real}, and a modified Laplacian derivative of luminance. This integration enables the generation of an illumination-corrected image optimized for improving the performance of camera-based modules as aforementioned in the previous work.


\vspace{-5pt}

\section{ The SENA Pipeline } 
\label{sec:validatingSimulation}


This section presents the proposed SENA pipeline designed to optimize the performance of camera-based modules. The full pipeline can be seen in Figure \ref{fig:follower_system}.

\vspace{-5pt}

\subsection{Preprocessing}


The images captured by the camera are initially in a gamma-compressed state, necessitating gamma decompression to facilitate the application of shadow removal techniques and achieve optimal results. Without this correction, shadowed regions in the image appear darker than their actual intensity. As illustrated in Figure \ref{fig:follower_system}, gamma correction enhances the visibility of details obscured within shadowed areas. Following gamma adjustment, the images are normalized to a range of 0 to 255, improving overall brightness. For the core processing, the YCbCr color space is utilized due to its ability to decouple luminance (Y), which represents image brightness, from chrominance (Cb and Cr), which encodes color information. This separation is particularly advantageous for shadow removal and related analyses.

\subsection{Shadow Detection}

Previous studies have demonstrated the effectiveness of shadow detection using the HSV color space, as illustrated in \cite{sabry2021drivable}. This paper incorporates the aforementioned approach to use the shadow region in the last stages of the SENA pipeline by utilizing the mean of the brightness channel V (Figure\ref{fig:follower_system}) as the upper threshold for identifying shadow regions. Specifically, the shadow region derived from the V channel, $S_{V}$, includes pixels whose values in channel V lie between 0 and $\mu_{V}$, where $\mu_{V}$ represents the mean brightness. Additionally, the luminance channel Y (Figure\ref{fig:follower_system}) from the YCbCr color space is employed to extract another shadow region, $S_{Y}$. Pixels in the Y channel with values less than or equal to 165 are classified as shadow pixels. The final shadow region, $S$ (Shadow Mask, Figure\ref{fig:follower_system}), is determined as the intersection of the two extracted shadow regions, $S_{V}$ and $S_{Y}$.


The shadow regions are used to refine the Y and Chroma channels with the red and blue color components, Cr and Cb respectively as seen in Figure \ref{fig:follower_system}.

\subsection{Illumination Invariant Channels}


To generate illumination-invariant images for AD applications, it is essential to employ plug-and-play methods that eliminate the need for preparatory steps, such as the entropy optimization required in logarithmic transformations, as described in \cite{shakeri2016illumination}. To achieve this, the SENA pipeline utilizes two distinct methods to create illumination-invariant images. The first method employs the mean chroma channel $m_c$, defined mathematically as $ m_c = (2A + Cb + Cr)/4$, where $A$ is the color component ranging from Green to Magenta from the LAB color-space, as seen in \cite{katramados2009real}.



The second approach is to use a normalized Laplacian 2nd derivative on the Y channel to get a more bright illumination invariant image, which use the approach in \cite{wang2007laplacian} as a baseline. 

The Laplacian of a 2D image is calculated by adding up the second x and y derivatives calculated using the Sobel operator in the \( x \)- and \( y \)-directions. 



To be able to use the Laplacian operator on a discrete 2D image, the Laplacian is computed by convolving a kernel \( \mathbf{K} \) with the image \( I(x, y) \) shown in Eq.\ref{eqn:lapacianFull}. This operator highlights regions of rapid intensity change, often used for edge detection.


\vspace{-10pt}

\begin{equation}
L(x, y) = \sum_{i=-1}^{1} \sum_{j=-1}^{1} K(i, j) \cdot I(x+i, y+j)
\label{eqn:lapacianFull}
\end{equation}

The resultant Laplacian image $L$ is then normalized from 0 to 255, which results in an image very close to a bright $m_c$ channel. 

\subsection{ Corrected Y Illumination Channel}
To correct the illumination channel, the first step is using the $L$ invariant channel. To return some edge features to $L$, contrast stretching is used. Contrast stretching is stretching intensity levels of image pixels. After multiple trials with different thresholds for contrast stretching, it was concluded that obtaining the pixel values at the 1.25 percentile and the 98.75 percentile and using them as the lower and upper limits of the contrast stretching operation showed sharpened details in the resultant image and restored some texture features. To compensate the Y channel, the mean value of the non-shadow pixels in the Y channel is normalized by 255. This mean is further used to normalize the inverse Y channel $Y^{norm}_{inv} = max(Y) - Y$. $Y^{norm}_{inv}$ is then used with $L$ to generate the illumination map for the Y channel $I^Y_{Map} = Y^{norm}_{inv} * L$. Shadow pixels in $I^Y_{Map}$ are then added to the original values in the Y channel as in Eq. \ref{eqn:finalY}. $Y_{out}$ is then normalized between 0 - 255.

\vspace{-10pt}

\begin{equation}
Y_{out}[shad] = Y[shad] + I^Y_{Map}[shad]
\label{eqn:finalY}
\end{equation}

\vspace{-5pt}

These methods generate a corrected illumination channel image that reduces shadows in sunny illumination conditions and enhances details in nighttime images with no need for any tunable parameters related to camera hardware or parameter optimization.

\vspace{-5pt}

\subsection{ Corrected Inverse Chromaticity Maps}

As mentioned in the introduction, varying shadow intensities in images present significant challenges. In order to mitigate them, the SENA pipeline uses inverse images of the chroma channels with some modifications to compensate each shadow pixel based on its intensity in both chromaticity channels accordingly. The original chromaticity channels Cb (Figure \ref{fig:follower_system}) $Ch^b$ with the blue color component and Cr (Figure \ref{fig:follower_system}) $Ch^r$ with the red color component, are inverted by subtracting the Max value from the full image. The resultant inverse image $Ch_{inv}$ is then normalized between 0 and 1 and used as a weighing matrix multiplied by $m_c$ resulting in $Ch_{Map}$, as depicted in Eq. \ref{eqn:inverse calculations}. The Chromaticity map $Ch_{Map}$ is then divided by the original chroma channel $Ch$ to generate the corrected inverse Chromaticity map $Ch^{Inv}_{Map}$. Finally, the $Ch^{Inv}_{Map}$ is subtracted from the original image to further brighten shadow regions $Ch^{Max}_{Inv}$. The final chroma channel corrections are different between $Ch^b$ and $Ch^r$, as they have shadow regions with inverse effects. In the $Ch^b$ channel, the shadows are darker than the rest of the image and in the $Ch^r$ channel, the shadows are brighter than the rest of the image. The corrected red chroma channel $Ch^r$ is formulated as the average of the $Ch_{Map}$, $Ch^{Max}_{Inv}$ multiplied by a Constant $C$ and the $m_c$. The corrected blue chroma channel $Ch^b$ is formulated by the average of $m_c$ and $Ch^{Inv}_{Map}$.



\begin{equation}
\begin{split}
    Ch_{inv} = max(Ch) - Ch, \qquad \quad\\
    Ch_{Map} = m_c * Ch_{inv}, Ch^{Inv}_{Map}=\frac{Ch_{Map}}{Ch}, \\
    Ch^r_{corrected} = \frac{Ch_{Map} + Ch^{Max}_{Inv} * C + m_c}{3}, \\
    Ch^b_{corrected} = \frac{m_c + Ch^{Inv}_{Map}}{2} \qquad \quad
\end{split}
\label{eqn:inverse calculations}
\end{equation}


In both chroma channels, the shadow regions are adjusted by multiplying the pixel values by the ratio of the corresponding pixels in the original channel to those in the modified chroma images. As a final fine-tuning step, the values of both chroma channels are normalized to match the range of the original channels prior to modification. This normalization minimizes the risk of significant color distortions in the image. Without this step, colors can become noticeably altered, for example, red may appear as bright orange or deep purple.

\section{ Experimental Setup }
\label{sec:ExperimentalResults}

This section presents the datasets and the evaluation metrics used to compare the SENA pipeline with CLAHE.

\vspace{-5pt}

\subsection{Datasets}

To be able to compare the results of the SENA pipeline with the CLAHE algorithm, 8 datasets were used, 5 from widely adopted datasets in CV and IP and three collected datasets within the AD context. An overview can be seen in Table \ref{tab:datasets}. The selected datasets demonstrate the SENA pipeline's capability in managing both broad and narrow scene perspectives under varying illumination conditions.

\vspace{-10pt}

\begin{table}[h]
\caption{Datasets Used for Evaluation}
\centering
\resizebox{0.9\columnwidth}{!}{
\begin{tabular}{|c|p{5.5cm}|}
\hline
\textbf{Dataset}                      & \textbf{Description}                                                                                      \\ \hline
\textbf{MEF}                          & 17 high-quality sequences (max size 384×512) covering diverse content, including natural scenes, indoor/outdoor views, and man-made structures \cite{ma2015perceptual}. \\ \hline
\textbf{LIME}                         & Challenging images with difficult illumination, excluding shadows in daylight \cite{guo2016lime}.  \\ \hline
\textbf{NPE}                          & 84 multi-scene natural images \cite{wang2013naturalness}.                                                 \\ \hline
\textbf{DICM}                         & 69 low-light images captured with commercial cameras \cite{lee2013contrast}.                             \\ \hline
\textbf{UNDD}                         & 75 dusk/night images from the unlabelled day and night dataset \cite{nag2019s}.                           \\ \hline
\textbf{Sunny Illumination}           & A set of 30 images that were used in \cite{sabry2021drivable}collected in sunny illumination conditions and pronounced shadows from a vehicle view.                                                    \\ \hline
\textbf{Night Driving}                & A set of over 33,000 frames from a dusk/nighttime road-driving video.                                              \\ \hline

\textbf{IAMCV Highway Samples}        & A set of 48 images with sunny illumination and mild shadows \cite{certad2024iamcv,certad2024iamcv2}. \\ \hline

\end{tabular} 
}

\label{tab:datasets}
\end{table}

\vspace{-10pt}

\begin{table*}[htp!]
\centering
\caption{The table presents the NR-IQA scores for CLAHE and the proposed SENA pipeline. The results indicate that the SENA pipeline outperforms CLAHE in four out of six metrics overall.}
\begin{tabular}{@{}l|c|cc|cc|cc@{}}
\toprule
\multicolumn{2}{c}{ } & \multicolumn{2}{|c|}{ \textbf{Average of All Datasets}  } & \multicolumn{2}{c|}{ \textbf{Night Driving Data} } &  \multicolumn{2}{c}{ \textbf{Sunny Illumination/Shadow Data} } \\
\toprule
\textbf{} & \textbf{Metrics} & \textbf{CLAHE} & \textbf{SENA Pipeline} & \textbf{CLAHE} & \textbf{SENA Pipeline} & \textbf{CLAHE} & \textbf{SENA Pipeline} \\
\midrule
\textbf{ } & $\downarrow$ Brisque & {\color{ForestGreen}36.56} & {\color{BrickRed}  85.34 }& {\color{ForestGreen} 47.11 } & {\color{BrickRed}  67.44 }& {\color{ForestGreen}42.65 } & {\color{BrickRed} 78.31}\\
\textbf{Lower Is Better} & $\downarrow$ NIQE    & {\color{BrickRed} 11.95 }& {\color{ForestGreen}10.37} & {\color{ForestGreen}15.1} & {\color{BrickRed} 16.99 }& {\color{BrickRed}12.06} & {\color{ForestGreen}6.97 }\\
\textbf{ } &$\downarrow$ PIQE    &{\color{BrickRed} 16.83} & {\color{ForestGreen}16.13} & {\color{BrickRed}23.29} & {\color{ForestGreen}10.38 } & {\color{BrickRed} 17.66 } & {\color{ForestGreen}12.02 } \\ \hline
\textbf{} & $\uparrow$ Mean Brightness   & {\color{BrickRed} 81.63 } & {\color{ForestGreen}161.62 }& {\color{BrickRed} 72.53  } & {\color{ForestGreen}166.32} & {\color{BrickRed} 140.65} & {\color{ForestGreen}195.54} \\
\textbf{Higher Is Better} & $\uparrow$ Standard Deviation      & {\color{ForestGreen}51.51} & {\color{BrickRed} 30.47 }& {\color{ForestGreen}50.65} & {\color{BrickRed} 22.92 }& {\color{ForestGreen} 51.69 } & {\color{BrickRed} 21.65}\\
\textbf{} & $\uparrow$ Entropy & {\color{BrickRed} 7.03}  & {\color{ForestGreen}7.35}  & {\color{BrickRed}6.88} & {\color{ForestGreen}12.39} & {\color{ForestGreen} 7.35 }& {\color{BrickRed} 5.87} \\
\bottomrule
\end{tabular}

\label{tab:dataset_res}
\end{table*}

To evaluate the SENA pipeline, No-Reference Image Quality Assessment (NR-IQA) metrics, also known as Blind Image Quality Assessment (BIQA), are employed \cite{anoop2024advancements}. In AD applications, it is challenging to capture identical images at different times of the day from a moving automated vehicle. Moreover, camera-based modules are required to operate in real time, necessitating direct no-reference evaluations. BIQA metrics evaluate image quality by analyzing natural scene statistics, with better scores reflecting images that are more visually appealing and align more closely with natural scene characteristics. The six BIQA metrics used in this paper consist of: 

\textbf{BRISQUE} (Blind/Referenceless Image Spatial Quality Evaluator), relies on natural scene statistics and does not need any distortion-specific knowledge. 

\textbf{Naturalness Image Quality Evaluator (NIQE)} uses a collection of quality-aware features based on a simple and effective natural scene statistic model. 

\textbf{Perception-based Image Quality Evaluator (PIQE)} focuses on blockiness, blurriness and noise. Inspired by principles of human perception of image quality

\textbf{Entropy} measures the amount of information or randomness in the image, indicating the level of detail/texture. 

Additionally, \textbf{Mean} and \textbf{Standard Deviation (SD)} provide basic statistical information about the brightness and contrast of images \cite{anoop2024advancements}.

To further evaluate the effectiveness of SENA compared to CLAHE as a preprocessing technique, a drivable area detection model, YOLOP as introduced by \cite{wu2022yolop}, was trained using a subset of nighttime driving images with challenging illumination conditions from the BDD100K dataset \cite{yu2020bdd100k}. These images preprocessed with CLAHE and SENA,  were then independently used to train the YOLOP network, focusing exclusively on the drivable area segmentation task. Two YOLOP networks were trained and tested with 990 nighttime images ( split between training, testing and validation ) for each preprocessing approach (SENA and CLAHE) for 50 epochs using an RTX 3090 GPU.

\vspace{-5pt}

\section{ Results }
\label{sec:results}
\subsection{Quantitative Results}

Table \ref{tab:dataset_res} presents results of the six NR-IQA metrics applied: (1) across all the datasets, (2) a nighttime scenario and (3) a sunny daytime illuminated environment with shadows. The results of the two approaches demonstrate that, despite the diverse and challenging illumination conditions, the SENA pipeline outperformed CLAHE in four out of the six metrics overall, namely: $PIQE$, $NIQE$, $Mean Brightness$ and $Entropy$. Significant improvements in perceptual quality and information can be seen particularly in poor lighting conditions at dusk or nighttime.

Regarding naturalness and visual perception quality, as measured by NIQE and PIQE respectively, the SENA pipeline surpassed CLAHE. The nighttime dataset revealed notable improvements in the PIQE metric, complemented by a substantial increase in the entropy score, nearly the double, which reflects greater amount of information and detail captured in images. The goal of enhancing images while reducing shadows resulted in the SENA approach achieving the highest average for Mean Brightness. The performance of the SENA pipeline was further statistically analyzed against CLAHE. The Wilcoxon signed-rank test was employed for all metrics due to the non-normal distribution of the data, as confirmed by the Shapiro-Wilk test. The results indicate statistically significant differences (p-value < 0.05) between the two methods for all metrics. Specifically, the SENA method demonstrated superior performance in terms of PIQE and Mean Brightness, suggesting improved perceptual quality and brighter image outputs. However, the Brisque and NIQE scores revealed potential trade-offs in naturalness and perceived quality, with higher values in the SENA method indicating possible degradation. Analysis of SD showed that the SENA method reduces contrast, leading to a more uniform intensity distribution. Furthermore, entropy analysis highlighted a reduction in information richness in the SENA method compared to CLAHE. These findings suggest that the SENA approach is effective in enhancing perceptual image quality but may involve trade-offs in contrast and detail preservation, depending on the application requirements. 


For the AD drivable area segmentation task, SENA outperformed CLAHE as a preprocessing step for YOLOP, as seen in Table \ref{tab:res_drivable_area}. This highlights the advantages of SENA in handling challenging nighttime driving scenarios, even with a limited amount of data samples. Figure \ref{fig:SENA_drivable} presents a sample output comparing the performance of CLAHE and SENA.
\vspace{-3pt}

\begin{table}[h!]
\centering
\caption{The scores for the drivable area segmentation using CLAHE, SENA and a Baseline without any preprocessing.}
\label{tab:metrics}
\begin{tabular}{lcc}
\toprule
\textbf{Approach} & \textbf{Accuracy} & \textbf{mIoU} \\
\midrule
Baseline & 0.758 & 0.550 \\
CLAHE  & 0.711 & 0.501 \\
SENA & \textbf{0.795} & \textbf{0.560} \\
\bottomrule
\end{tabular}
\label{tab:res_drivable_area}
\end{table}

\vspace{-5pt}


\subsection{Qualitative Results}

\begin{figure}[t]
\centering
  \includegraphics[width=0.9\linewidth]{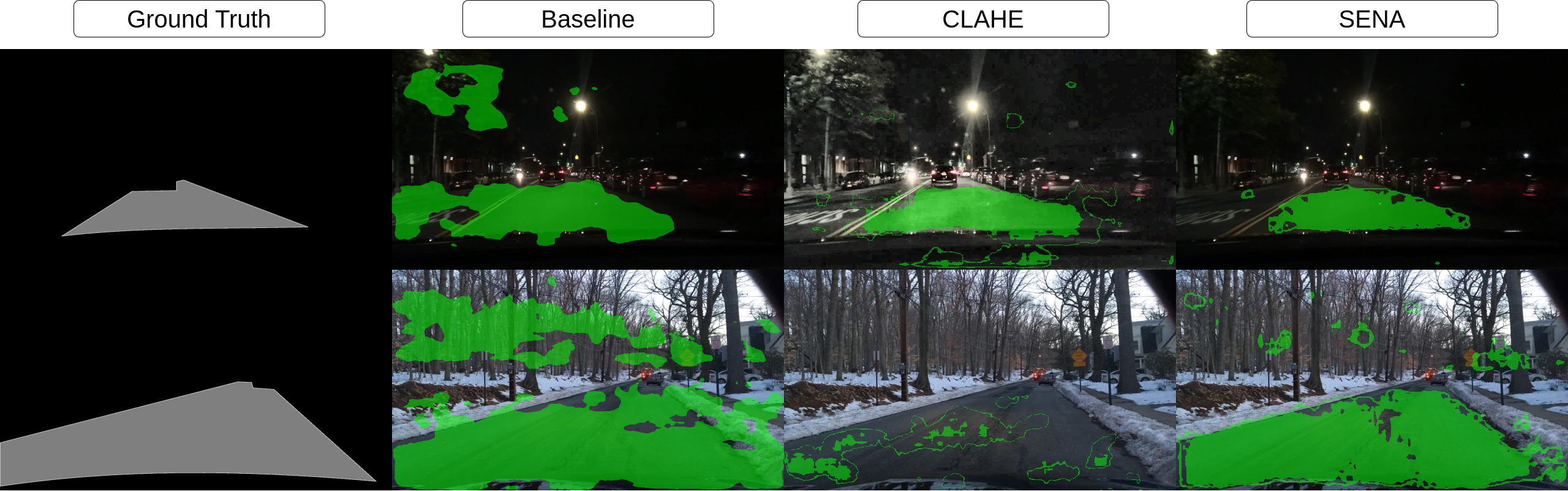}
\caption{A sample output of drivable area detection using YOLOP demonstrates the application of CLAHE, SENA and the Baseline without any preprocessing. Despite the dataset being deliberately curated to include a significant proportion of nighttime samples, SENA exhibited robust performance in daytime conditions as well, surpassing CLAHE and being more precise than the Baseline. } 
\label{fig:SENA_drivable}
\end{figure}

\begin{figure}[t]
\centering
  \includegraphics[width=0.7\linewidth]{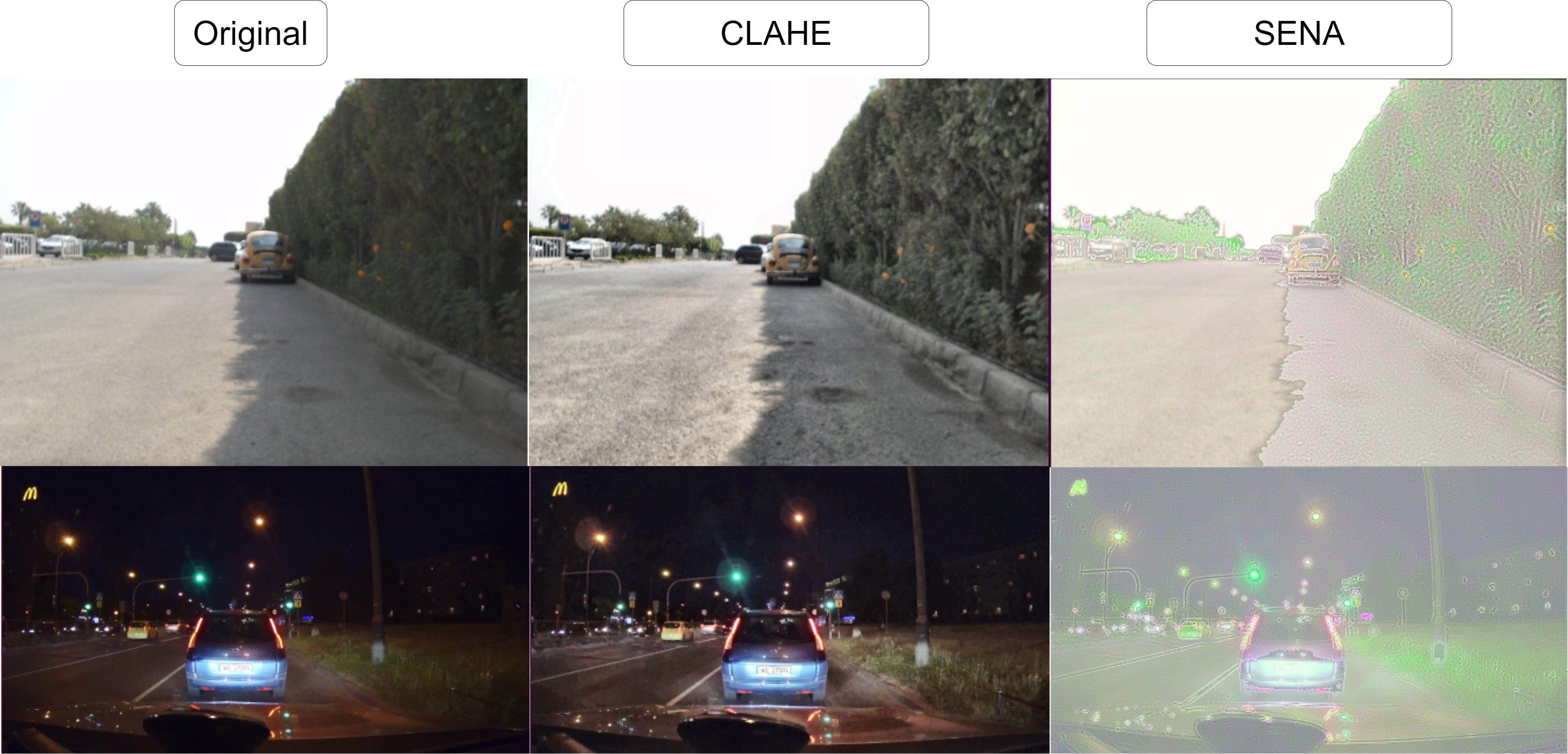}
\caption{An example output from CLAHE and the SENA approach. With the main target to compensate shadows in daytime images, and make poorly visible objects more visible, the SENA approach is capable of addressing image enhancement in both tasks. The top row presents a day time image with shadows. The bottom row presents a nighttime scenario. }\label{fig:example_out}
\end{figure}

In this section, samples of output images from CLAHE and the SENA approach are visually assessed, as shown in Figure \ref{fig:example_out}. The SENA approach successfully achieves the goals of shadow correction and nighttime image enhancement. CLAHE, on the other hand, produces sharper images but does not compensate shadow regions or brighten dark features at nighttime. Overall, the SENA approach balances shadow correction and nighttime enhancement while preserving naturalness.

For drivable area detection, using SENA as a preprocessing step outperformed CLAHE in both Accuracy and Mean Intersection over Union (mIoU). Although the dataset was manually curated from the BDD100K dataset to predominantly include nighttime images, it contains very few daytime images—fewer than 20 out of 990 samples. This typically leads to class imbalance and is expected to result in poor segmentation performance for daytime images, as demonstrated by the output of CLAHE in Figure \ref{fig:SENA_drivable}. However, YOLOP trained with SENA was able to adapt to some daytime images, suggesting that SENA preserves consistent features across varying illumination conditions, thereby improving performance under both nighttime and daytime scenarios.

\section{Conclusion and Future Work}
\label{sec:Conclusions}
\vspace{-5pt}

Based on the scores presented in Table \ref{tab:dataset_res} and Table \ref{tab:res_drivable_area}, alongside the visual comparisons in Figure \ref{fig:example_out} and Figure \ref{fig:SENA_drivable}, the SENA approach demonstrates superior performance in terms of naturalness and perceptual quality compared to CLAHE. Additionally, SENA achieves better results for drivable area detection when compared to CLAHE.

The primary focus of this work was to achieve general enhancement for both shadow mitigation in daytime images and nighttime image enhancement in one step. In addition, drivable area segmentation during nighttime conditions was utilized to highlight the advantages of the proposed SENA approach. 
The SENA method resulted in a suitable preprocessing step for camera-based modules in DL or classical approaches, including drivable area segmentation. As mentioned in Section \ref{sec:Introduction}, one advantage of the SENA approach is that it requires no tuning. Additionally, in contrast to \cite{maddern2014illumination}, this approach does not depend on any parameter specific to the image sensor. Given the performance of the model as discussed in Section \ref{sec:results}, it could be employed as a pre-processor for the classical tasks of AD.

It is also worth noting that this approach fares well with the Perception-based Image Quality Evaluator (PIQE) indicating that the resulting images are appropriate for human monitoring. As mentioned in \cite{higashiyama2024investigation}, PIQE is inspired by principles of human perception of image quality. As AVs come into daily operation, remote operators may be required to ensure that the AVs operate safely in the real-world. 


In conclusion, SENA outperforms CLAHE in the PIQE metric across daytime, nighttime, and scenarios with non-uniform illumination, resulting in improved overall visibility under challenging lighting conditions. By generating illumination-corrected images without requiring tuning or optimization, while preserving color information, the algorithm shows immense potential for integration into other AD applications given its performance in drivable area segmentation. Additionally, SENA operates at 25 ms for 640×480 images in Python programming language using a i7-11700H laptop CPU, making it suitable for real-time use.

Future work will focus on integrating the proposed approach as a preprocessing step for camera-based modules, such as object detection and segmentation. Furthermore, it will be tested within a complete AD software stack to evaluate its impact on overall system performance.







\vspace{-5pt}

\section*{Acknowledgment}

This work was funded by the Austrian Research Promotion Agency (FFG), PDrive, project number: 12451001.

\bibliographystyle{IEEEtran}
\bibliography{paper}






\vspace{-5pt}


\end{document}